\def\year{2022}\relax
\documentclass[letterpaper]{article} 
\usepackage{aaai22}  
\usepackage{times}  
\usepackage{helvet}  
\usepackage{courier}  
\usepackage[hyphens]{url}  
\usepackage{graphicx} 
\usepackage{natbib}  
\usepackage{caption} 
\DeclareCaptionStyle{ruled}{labelfont=normalfont,labelsep=colon,strut=off} 
\usepackage{algorithm}
\usepackage{algorithmic}

\usepackage{newfloat}
\usepackage{listings}
\floatstyle{ruled}
\newfloat{listing}{tb}{lst}{}
\floatname{listing}{Listing}
\usepackage{pgfplots}
\pgfplotsset{compat=1.11,every axis/.append style={tick label style={font=\tiny}},
    /pgfplots/ybar legend/.style={
    /pgfplots/legend image code/.code={%
       \draw[##1,/tikz/.cd,yshift=-0.25em]
        (0cm,0cm) rectangle (3pt,0.8em);},
   },
}
\usepackage{multirow}
\usepackage{xcolor, colortbl}
\definecolor{_blue}{HTML}{4168E2}
\definecolor{_red}{HTML}{E01051}
\definecolor{customred}{HTML}{E01051}
\definecolor{_yellow}{HTML}{F0DF9C}
\usepackage{comment}

\usepackage{multirow}

\newcommand{\locationtype}[2]{%
 \begin{tikzpicture}[every node/.style={inner sep=2,outer sep=0}]
	\node[draw,fill=#1]{\tiny#2};
\end{tikzpicture} 
}
\newcommand{\locationentity}[2]{%
 \begin{tikzpicture}[every node/.style={inner sep=1.5,outer sep=0}]
	\node[draw,fill=#1]{#2};
\end{tikzpicture} 
}
\newcommand{\locationtag}[2][\relax]{\locationentity{red!25!white}{#2}$_{\ifx#1\relax\bigovoid\else\locationtype{blue!15!white}{#1}\fi}$}
\title{Identification of Fine-Grained Location Mentions in Crisis Tweets}
\author {
    Sarthak Khanal,
    Maria Traskowsky, 
    Doina Caragea 
}
\affiliations {
     Kansas State University\\
    sarthakk@ksu.edu, mariatraskowsky@ksu.edu, dcaragea@ksu.edu
}

\usepackage{bibentry}

\begin{document}

\maketitle

\begin{abstract}
Identification of fine-grained location mentions in crisis tweets is central in transforming situational awareness information extracted from social media into actionable information. Most prior works have focused on identifying generic locations, without considering their specific types. To facilitate progress on the fine-grained location identification task, we assemble two tweet crisis datasets and manually annotate them with specific location types. The first dataset contains tweets from a mixed set of crisis events, while the second dataset contains tweets from the global COVID-19 pandemic. We investigate the performance of  state-of-the-art deep learning models for sequence tagging on these datasets, in both in-domain and cross-domain settings.
\end{abstract}

\section{Introduction}
We have witnessed a large number of crisis situations in recent years, from natural disasters to man-made disasters and also to deadly animal and human health crises, culminating with the ongoing COVID-19 public health crisis. 
Affected individuals often turn to social media (e.g., Twitter or Facebook) to report useful information, 
or ask for help 
\cite{sakaki2010earthquake,vieweg2010microblogging,king2018social}. Information contributed on social media by people on the ground can be invaluable to emergency response organizations in terms of gaining situational awareness,  prioritizing  resources to best assist the affected population, addressing concerns, and even saving lives \cite{king2018social}.

\begin{table*}[ht]
\scriptsize
\centering
\begin{tabular}{|c|l|}
\hline
\multicolumn{1}{|l|}{{\bf No.}} & {\bf Tweet text}                                                                                                      \\ \hline

 \multirow{2}{*}{1}        & 
Roads in {\colorbox{red!25!white}{Calhoun County}} are underwater, access to the {\colorbox{red!25!white}{Port Lavaca Causeway}} is flooded, the bridge is closed. \\ \cline{2-2} 
                          &      
                  {\color{white}{oo}}O {\color{white}o} 
                  O{\color{white}}  {\colorbox{blue!25!white}{B-ctc{\color{blue!25!white}OO} I-ctc {\color{blue!25!white}oi}}} {\color{white}o}O{\color{white}ooooo}O{\color{white}oooooo} O{\color{white}o}O{\color{white}o}O{\color{white}oi}{\colorbox{blue!25!white}{B-lan {\color{blue!25!white}i}I-lan {\color{blue!25!white}ooo} I-lan{\color{blue!25!white}o}}}{\color{white}{o}}O{\color{white}{oo}}O{\color{white}{oooo}}O{\color{white}{ooo}}O{\color{white}{oo}}O{\color{white}{oo}}O
                          \\
                          \hline 
 \multirow{2}{*}{2}        & 
Very extensive damage sustained throughout {\colorbox{red!25!white}{Wilmington}}  {\colorbox{red!25!white}{NC {\color{red!25!white}o}}} from Hurricane Florence  \\ \cline{2-2} 
                          &      
                  {\color{white}{oo}}O {\color{white}ooo} 
                  O{\color{white}ooooo} O{\color{white}ooooo} O{\color{white}ooooooo}
                  O{\color{white}oooo}
             {\colorbox{blue!25!white}{\hspace{0.3cm} B-ctc  {\color{blue!25!white}oo}}} {\color{white}}{\colorbox{blue!25!white}{{\color{blue!25!white}}B-sta{\color{blue!25!white}}}}{\color{white}I} {\color{white}o}
             O{\color{white}oooo}
              O{\color{white}ooooo}
               O{\color{white}ooooo}
                          \\ \hline             
                          
  \multirow{2}{*}{3}        & 
Big tree fell on power lines and blocking {\colorbox{red!25!white}{Brown Ave}} near {\colorbox{red!25!white}{Washington St}} in {\colorbox{red!25!white}{Orlando}} s {\colorbox{red!25!white}{Thornton Park}}  \\ \cline{2-2} 
                          &      
                  {\color{white}{o}}O {\color{white}o} 
                  O{\color{white}o} O{\color{white}o} O{\color{white}oo}
                  O{\color{white}ooo}
                   O{\color{white}oo}
                    O{\color{white}ooo}
                     O{\color{white}ooo}
                  {\colorbox{blue!25!white}{B-oth I-oth{\color{blue!25!white}}}}{\color{white}o} O{\color{white}ii} {\color{white}}{\colorbox{blue!25!white}{{\color{blue!25!white}O}B-oth I-oth{\color{blue!25!white}O}}} O{\color{white}}{\color{white}i}{\colorbox{blue!25!white}{B-sta\hspace{0.3cm}}}{\color{white}i}O{\color{white}} {\colorbox{blue!25!white}{B-lan I-lan{\color{blue!25!white}ooi}}}
                          \\ \hline                             

\multirow{2}{*}{4}        &There are now more confirmed cases of coronavirus in {\colorbox{red!25!white}{New {\color{red!25!white}i} York {\color{red!25!white}i} City}} than there are in all of {\colorbox{red!25!white}{South Korea}} 
   \\ \cline{2-2} 
                          &      
                  {\color{white}{oi}}O{\color{white}oo}
                  O{\color{white}ii} O{\color{white}ooo} O {\color{white}{oooo}}O {\color{white}{ooooo}}O {\color{white}{oi}}O {\color{white}{oooo}}O {\color{white}{oooii}}O {\colorbox{blue!25!white}{B-ctc I-ctc \hspace{.1cm} I-ctc }}{\color{white}oo}O{\color{white}ooo}O{\color{white}o} O{\color{white}o}O{\color{white}o}O {\color{white}i}O{\color{white}i}{\colorbox{blue!25!white}{B-con I-con}}
                          \\ \hline
\multirow{2}{*}{5}        & {\colorbox{red!25!white}{South Asia \hspace{.1cm}}} is quickly marching towards being the new epicenter of covid 19  \\ \cline{2-2} 
                          &      
                          {\colorbox{blue!25!white}{B-reg {\color{blue!25!white}i}I-reg }}
                  {\color{white}{}}O {\color{white}oo} 
                  O{\color{white}ooooo} O{\color{white}ooooo} O{\color{white}oooo}  O{\color{white}oo}
                  O{\color{white}oo}
                  O{\color{white}oooo}
                   O{\color{white}ooo}
                    O{\color{white}o}
                     O{\color{white}oo}
                      O{\color{white}}
                          \\ \hline  
\multirow{2}{*}{6}        & The difference in COVID 19 cases and deaths between {\colorbox{red!25!white}{New York}} and {\colorbox{red!25!white}{California}} continues to be astounding 
   \\ \cline{2-2} 
                          &      
                  {\color{white}{o}}O{\color{white}oooo}
                  O{\color{white}ooo} O{\color{white}ooi} O {\color{white}{ooo}}O {\color{white}{oo}}O {\color{white}{oi}}O {\color{white}{ooi}}O {\color{white}{ooooi}}O
                  {\color{white}{oo}}
                  {\colorbox{blue!25!white}{B-sta I-sta}}{\color{white}o}O{\color{white}ii} {\colorbox{blue!25!white}{\hspace{.2cm}B-sta \hspace{.2cm}}}{\color{white}oooo}O{\color{white}ooo} O{\color{white}i}O{\color{white}iooo}O {\color{white}}
                          \\ \hline

\end{tabular}
\caption{Examples of crisis tweets tagged with fine-grained location types. The subsequences representing location mentions are highlighted with pink, and their corresponding tags (in BIO format) are highlighted with blue.}
\label{examples}
\end{table*}
Many recent studies have focused on identifying informative tweets posted by  individuals affected by a crisis, and classifying those tweets according to situational awareness categories useful for crisis response and management \cite{imran2015processing}. However, for situational awareness information extracted from social media to be actionable,  knowing the corresponding geographic location is of key importance. For example, location information enables responders to perform fast assessment of the damage produced by a natural disaster \cite{villegas2018lessons}, or to respond to requests for help coming from affected individuals or institutions (e.g., hospitals or schools).  In the case of COVID-19 health crisis, location information can also be used to identify trends by locations (e.g., stance of a community towards various health recommendations) \cite{mutlu2020stance, miao-etal-2020-twitter}, and subsequently employ that information to prevent dissemination of misinformation and rumors, and resurgence of the novel coronavirus.

Unfortunately, only a very small percentage of tweets are geotagged \citep{ICWSM124605}. Furthermore, even when  geolocation information is available, that  location may not be the location mentioned in the tweet text \citep{ikawa2013location}. According to \citet{vieweg2010microblogging}, the location in the tweet text is usually the location needed for monitoring and/or responding to an emergency. Table \ref{examples} shows several examples of tweets posted during recent hurricanes (first three tweets) and during the COVID-19 crisis (last three tweets). As can be seen, locations are mentioned at different levels of granularity, from  region and landmark to city, state and country. Furthermore, the same location name, in our COVID-19 examples - {\it New York}, can be associated with different location types, such as {\it city} (tweet 4) and  {\it state}  (tweet 6). Information about the tags of the ambiguous entities can be used to disambiguate the corresponding locations and link them to physical locations. Therefore,  tools for identifying fine-grained locations  directly from the texts of crisis tweets are greatly needed. 

Location identification has been frequently addressed as part of the broader named entity recognition (NER) task \citep{goyal2018recent,li2020survey}. Some studies have focused specifically on the task of identifying generic location mentions (without considering the type of location) in tweet text \citep{hoang2017predicting}, and even disaster tweet text \citep{kumar2019location}.  
Other studies have focused on identifying  fine-grained points-of-interest (POI), useful for location-based services \citep{li2014fine,malmasi2015location,ji2016joint,xu2019dlocrl}. 

To the best of our knowledge, there are no publicly available, manually annotated datasets that can facilitate progress  on the task of identifying fine-grained locations (including, city, state, country, region, landmark) in crisis tweets, despite the benefits provided by the use of social media data in  monitoring and responding to a crisis. 
To address this need, we have assembled two datasets for identifying fine-grained locations in crisis tweets.  The first dataset, called {\it MIXED}, consists of tweets crawled during five crisis events, specifically, Nepal Earthquake, Queensland Floods, Srilanka Bombing, Hurricane Michael and Hurricane Florence. The second dataset, called {\it COVID}, consists of a set of coronavirus-related tweets crawled between February 27 and April 7, 2020. We used Amazon Mechanical Turk (AMT)\footnote{\url{https://www.mturk.com/}} to annotate the datasets using six location types (Country, State, Region, City, Landmark and Others).

Given the success of deep learning approaches for NER tasks \citep{li2020survey},  we use different state-of-the-art models to establish baseline results on the dataset. 
 In summary, the contributions of this work are as follows: 

\begin{itemize}
    \item We create two datasets of tweets from a mixed set of crisis events and from   COVID-19, respectively. The tweets are manually annotated with fine-grained location types, including city, state, country, region, landmark.
    \item {We use state-of-the-art models including a contextual encoder coupled with a tag decoder in a multi-task learning setting, and a model based on contextualized word and entity representations, combined with entity-aware self-attention to establish baseline results for our datasets.}
    \item We perform extensive experiments on the {\it MIXED} and {\it COVID} datasets, respectively, in  both {\it in-domain} and {\it cross-domain settings} to understand the usefulness of the data from the domain of interest, as well as the transferability of the models from one domain to another.
\end{itemize}
Given this introduction, we proceed with a discussion of related work in the next section, followed by the description of the datasets constructed, and then  background and approaches, experimental setup, results and error analysis, and finally, conclusions and an ethics statement. 
\section{Related Work}
\label{related}
We organize the related work based on several categories relevant to the research in this paper. Specifically, we first briefly discuss the location mention identification as a specific task in the area of NER. Subsequently, we review works on fine-grained location types, followed by approaches used for identifying locations, and finally, other existing and relevant location datasets.

\subsection{NER and location mention identification} NER is a well-researched problem in natural language processing (NLP) \citep{goyal2018recent,li2020survey}. 
Text-based location identification has been traditionally addressed as part of the broader NER task,  although some works focus specifically on location identification \citep{lingad-location,han2014identifying,kumar2019location,magge2019bi}.
Most of the works that identify locations simply tag location mentions, as opposed to identifying fine-grained location types \citep{li2020survey}. For example, 
\citet{lingad-location} aim to identify mentions of locations (including geographic locations and points of interest) in disaster tweets, by using standard  NER taggers (pre-trained or retrained), and report best performance using retrained Stanford NER \citep{Finkel:2005:INI:1219840.1219885}.
Also in the context of emergencies,  \citet{kumar2019location} use a convolutional neural network (CNN) approach to identify location references
in crisis tweets, regardless of their  specific types.

\subsection{Fine-grained location types} Some recent works have considered fine-grained location types, such as city, state, country \citep{10.1007/978-3-319-18117-2_24,anand2017fine,lal2019sane,qazi2020geocov19}. While focused on COVID-19 tweets, \citet{qazi2020geocov19} use a gazetteer approach to infer the geolocation of tweets, based on user and tweet information.  Closest to our goal of identifying fine-grained locations in disaster tweet texts, \citet{10.1007/978-3-319-18117-2_24} propose a CRF-based approach to identify countries, states/provinces and cities using a Twitter dataset annotated according to guidelines provided in \citep{10.2307/40666361}. They make use of hand-crafted features, including gazetteer features, to train a CRF model. As opposed to \citep{10.1007/978-3-319-18117-2_24}, we use a larger set of location types and approaches that preclude the need for manually crafted features and  gazetteers. 

 Other works on fine-grained location focus on  identifying point of interests locations, such as restaurants, hotels, parks, etc. and linking them to pre-defined location profiles \citep{li2014fine,ji2016joint,xu2019dlocrl}. 
 \citet{li2014fine} build a POI inventory (which can be seen as a noisy version of a gazetteer), and  a time-aware POI tagger. The time-aware POI tagger is a CRF trained to extract and disambiguate fine-grained POIs. 
 \citet{ji2016joint} extend the POI tagger in  \citet{li2014fine} by proposing a joint framework that achieves POI recognition and linking to  pre-defined POI profiles simultaneously.   \citet{xu2019dlocrl}  address the same problem of identifying fine-grained POIs and linking them to location profiles. However, they use a deep learning model (specifically, BiLSTM-CRF) to avoid the need for manually designed features, and subsequently use a collection of location profiles to perform the linking. The definition of fine-grained POI tagging is different from our definition of fine-grained location tagging - we aim to assign specific types/tags to location entities, as opposed to identifying generic (yes/no) POI tags, and then linking the tags to pre-defined profiles, as  in prior works \citep{li2014fine,ji2016joint,xu2019dlocrl}. Moreover, we want to avoid the use of gazetteers to ensure that the models are resilient to the informal nature of the language used in tweets. Similar to \citep{xu2019dlocrl}, we also want to avoid the need for manually designed features, and  thus focus on deep learning approaches. 

\subsection{State-of-the-art  approaches for NER} State-of-the-art approaches for NER, in general,  and location identification, in particular, are sequence labeling type approaches based on deep learning language models  \citep{li2020survey}. More specifically, competitive  architectures consist of three components: distributed representations of the input, a context encoder model, and a tag decoder model
. Both character-level and word-level embeddings (or their combination)  have been used to represent the NER input in recent works \cite{goyal2018recent}, with BERT \citep{DBLP:journals/corr/abs-1810-04805}  contextual embeddings being among the most successful \cite{li2020survey}. In terms of context encoders and tag decoders, recurrent neural networks, most often, BiLSTM networks (short for Bidirectional Long Short-Term Memory) \citep{hochreiter1997long}, and CRF (short for Conditional Random Fields) \citep{lafferty2001conditional}, respectively, contribute to  some of the best results on benchmark NER datasets \citep{luo2019hierarchical,baevski2019cloze,liu2019gcdt,jiang-etal-2019-improved}. Given these successful architectures for the NER task, one of our baseline models consists of three components: BERT, BiLSTM and CRF, for the input representation, context encoder and tag decoder, respectively. As another strong baseline, we investigate a recent state-of-the-art architecture, called LUKE, \citep{yamada-etal-2020-luke}, based on a bidirectional transformer architecture pre-trained to output both word and entity contextualized representations. LUKE uses an entity-aware self-attention to identify entities. 

\subsection{Existing location datasets}
Most previous works on location identification in tweet texts are focused on general tweets \cite{liu2014automatic,10.1007/978-3-319-18117-2_24} with a few notable exceptions of works focused on crisis tweets \citep{lingad-location,kumar2019location,qazi2020geocov19}. However, the datasets used in these works are not all available  \citep{lingad-location,kumar2019location}. Even when available,  the datasets focus on identifying location mentions without specifically identifying the fine-grained type of the location mentions \citep{liu2014automatic}. \citet{qazi2020geocov19} used a gazetteer-only approach to annotate tweets with geolocations, and the resulting annotations are not very accurate. While not specifically focused on crisis tweets, the dataset published by \citet{10.1007/978-3-319-18117-2_24} is the closest to our dataset in terms of fine-grained location types used (which include city, country, state or province, etc.). However, most locations in their  dataset are not mentioned in the tweet, but are inferred from auxiliary information. Specifically, only about 3\% of the tweet texts in their dataset have location entities, for a total of only 220 different location entities. Furthermore, they also used a gazetteer approach to annotate most of the tweets, and performed  manual annotations just for a small subset of their dataset. Given the above-mentioned differences between existing datasets and our datasets, it is not possible to directly use the existing datasets to transfer information to our tasks in a cross-domain setting.

\section{Datasets}
\label{datasets}
One main contribution of our work is to construct two benchmark datasets for identifying fine-grained locations (see Table \ref{tab:category_distribution}) useful for crisis monitoring and response. The datasets cover events that are different in nature, to enable studies in both in-domain and cross-domain settings.

\subsection{Data collection} The first dataset, called {\it MIXED}, contains tweets posted during four natural disasters and one man-made disaster that happened in specific geographical regions. The second dataset, called {\it COVID}, contains tweets posted during the COVID-19 pandemic, and thus has worldwide coverage. More specifically, the tweets in the {\it MIXED} dataset were crawled during the following events: Nepal Earthquake, Queensland Floods, Srilanka Bombing, Hurricane Michael and Hurricane Florence. The tweets from Nepal Earthquake and Queensland Floods were obtained from \citep{alam-etal-2018-domain}. Tweets from Srilanka Bombing, Hurricane Michael and Hurricane Florence were crawled locally using the Twitter streaming API. A random sample of unique English tweets was included in the  {\it MIXED} dataset that was annotated using  AM. More than 133 million tweets from COVID-19 pandemic  were also crawled locally between February 27th and April 7th, 2020. A random sample of unique English tweets was included in the  {\it COVID} dataset for AMT annotation. The keywords used to crawl the tweets and the final number of tweets included in the dataset for each event are provided in the appendix

{\it Unlabaled tweets.} In addition to the {\it MIXED} and {\it COVID} datasets that are annotated as part of this work, we also used a large number of unlabeled mixed crisis and COVID-19 tweets to pre-trained BERT \cite{DBLP:journals/corr/abs-1810-04805} models and obtain crisis-specific embeddings. In particular, to pre-train the BERT model for the \textit{MIXED} dataset, we collected a larger set of tweets pertaining to various crisis events from prior works \citep{imran2016lrec,nguyen2017robust,crisismmd2018icwsm,alam-etal-2018-domain,olteanu2014crisislex,Olteanu:2015:EUH:2675133.2675242} in addition to the locally crawled tweets. For the \textit{COVID} dataset, however, we only used the locally crawled tweets to pre-train the BERT model.

\begin{table}[t]
\scriptsize
\centering
\begin{tabular}{|l|l|rr|rr|}
\hline
\multicolumn{1}{|l|}{\multirow{2}{*}{\textbf{Type}}} & \multirow{2}{*}{\textbf{Descr.}}                                           & \multicolumn{2}{c|}{\textbf{MIXED Distr.}}                              & \multicolumn{2}{c|}{\textbf{COVID Distr.}}                          \\
\multicolumn{1}{|l|}{}                      &                                                                        & \multicolumn{1}{r}{\textbf{\#}}   & \multicolumn{1}{c|}{\textbf{\%}}     & \multicolumn{1}{r}{\textbf{\#}}   & \multicolumn{1}{c|}{\textbf{\%}} \\ \hline
con                                         & Country                                                                & 1,763                     & 28.31                      & 1,819                     & 53.06                  \\ \hline
sta                                         & State                                                                  & 1,242                       & 19.95                      & 396                       & 11.56                  \\ \hline
reg                                         & \begin{tabular}[c]{@{}l@{}}Region, \\ Continent\end{tabular}         & 764                     & 12.27                      & 158                       & 4.61                  \\ \hline
ctc                                         & \begin{tabular}[c]{@{}l@{}}City, Town,\\ County\end{tabular}      & 797                       & 12.80                      & 518                       & 15.11                   \\ \hline
lan                                         & \begin{tabular}[c]{@{}l@{}}Building,\\Landmark \end{tabular}        & 1,190                     & 19.11                      & 391                       & 11.42                  \\ \hline
oth                                         & \begin{tabular}[c]{@{}l@{}}Other  \\ \end{tabular} & 471                       & 7.56                       & 146                       & 4.24                   \\
\hline
\multicolumn{1}{|l|}{All}                   & Entities                                                               & \multicolumn{1}{r}{6,227} & \multicolumn{1}{r|}{100.00} & \multicolumn{1}{r}{3,428} & 100.00  \\
\hline
\end{tabular}
\caption{Location types and their descriptions, together with type distribution (as raw numbers \# and percentages \%) in the {\it MIXED} and {\it COVID} datasets, respectively.}
\label{tab:category_distribution}
\end{table}

\subsection{Data annotation and quality assessment}
To prepare the tweets for annotation, the following pre-processing was performed. User mentions were anonymized by replacing them with a generic {\tt user} keyword, and links were removed from the tweet text. Special characters, including -=~!\#\$\% $\hat{ }$ \&*()+[]\{\};\textbackslash':"$\mid$$<$$>$?, and non-printable ASCII characters 
were also removed. The tweet text was tokenized 
to enforce annotation at the token level and avoid accidental annotation of token fragments. Tweet  tokens were annotated with six location types using the BIO scheme (where B stands for Beginning, I stands for Inside and O stands for Outside of a location entity).  The location types together with their brief descriptions are shown in Table \ref{tab:category_distribution}. Examples of annotated tweets are shown in Table \ref{examples}, where the first three tweets are representative of the {\it MIXED} dataset, and the last three are representative {\it COVID}. 

We used feedback from a local annotator to iteratively develop and improve a custom annotation tool for our task. The tool was subsequently deployed to AMT. 
Annotators were provided with definitions of the location types included in our study, together with precise instructions for annotation, and examples of annotated tweets, such as those in Table \ref{examples}. 
Each tweet was annotated by at least 3 annotators. Only  entities where two or more annotators agreed were included in the final datasets. 
The Cohen's Kappa  scores that we obtained for inter-annotator agreement were 0.63 and 0.62, and the average pairwise F1-scores for inter-annotator agreement were 68.87 and 65.86 for the {\it MIXED} and {\it COVID} datasets, respectively. According to \citet{doi:10.1177/001316446002000104}, these scores represent substantial agreement. 

The distributions of the location entities over the six location types included in our study are shown in Table \ref{tab:category_distribution}, 
As can be seen, the annotated entities are more evenly distributed over the types considered in the {\it MIXED} dataset, while  more than half of the entities are of type {\it country} in the {\it COVID} dataset. The datasets also show differences in terms of the number of entities per tweet, with the {\it MIXED} dataset containing a majority of tweets with one or two entities (and a small number of tweets with more than two entities), and  {\it COVID}  containing mostly tweets with one entity (and a small number of tweets with two or more entities). Such differences emphasize  specific characteristics and challenges in the two domains, and are useful in studying the transferability of the models from one domain to another. 
\begin{table}[t]
\scriptsize
\centering
\begin{tabular}{|l|l|r|r|r|r|}
\hline
\textbf{Dataset}                 & \textbf{No.}      & \multicolumn{1}{l}{\textbf{Train}} & \multicolumn{1}{|l|}{\textbf{Test}} & \multicolumn{1}{l}{\textbf{Dev}} & \multicolumn{1}{|l|}{\textbf{Total}} \\ \hline
\multicolumn{1}{|c|}{}      & Tweets   & 2,620                     & 820                      & 656                     & 4,096                     \\
\multicolumn{1}{|c|}{} & Tokens   & 73,622                    & 23,253                   & 18,511                  & 115,386                   \\
\multicolumn{1}{|c|}{}      & Entities & 4,001                     & 1,237                    & 989                     & 6,227\\
\cline{2-6}
\multicolumn{1}{|l|}{}                                  & \multicolumn{5}{c|}{Entity type distribution}      \\
\cline{2-6}
\multicolumn{1}{|c|}{}      & con & 1,135                    & 360                    & 268   & 1,763\\ 
 MIXED                      & sta & 752                     & 267                    & 223   & 1,242\\
\multicolumn{1}{|c|}{}      & reg & 514                     & 144                    & 106   & 764\\
\multicolumn{1}{|c|}{}      & ctc & 522                     & 141                    & 134   & 797\\
\multicolumn{1}{|c|}{}      & lan & 768                     & 238                    & 184   & 1,190\\
\multicolumn{1}{|c|}{}      & oth & 310                     & 87                     & 74    & 471\\
\hline
\hline
                          & Tweets   & 3,355                     & 1,049                    & 839                     & 5,243                     \\
\multicolumn{1}{|c|}{}                     & Tokens   & 103,646                   & 32,674                   & 25,798                  & 162,118                   \\
                          & Entities & 2,206                     & 656                      & 566                     & 3,428             \\
\cline{2-6}
\multicolumn{1}{|l|}{}                                  & \multicolumn{5}{c|}{Entity type distribution}      \\
\cline{2-6}
\multicolumn{1}{|c|}{}      & con & 1,162                    & 347                   & 310  & 1,819\\ 
 COVID                      & sta & 264                     & 75                    & 57   & 396\\
\multicolumn{1}{|c|}{}      & reg & 101                     & 34                    & 23   & 158\\
\multicolumn{1}{|c|}{}      & ctc & 328                     & 103                   & 87   & 518\\
\multicolumn{1}{|c|}{}      & lan & 265                     & 62                    & 64   & 391\\
\multicolumn{1}{|c|}{}      & oth & 86                      & 35                    & 27   & 146\\
                          \hline
\end{tabular}
\caption{Statistics for the number of tweets, tokens and the number of location entities in the {\tt train/test/dev} subsets of the {\it MIXED} and {\it COVID} datasets, respectively. The entity type distribution in the {\tt train/test/dev} subsets  is also shown for each dataset.}
\label{train-test-dev}
\end{table}

\subsection{Benchmark Datasets} To enable progress on fine-grained location identification in crisis tweets, and facilitate comparisons between models developed for this task (in-domain and cross-domain), we created benchmark datasets  by randomly splitting our {\it MIXED} and {\it COVID} datasets into training ({\tt train}), development ({\tt dev}) and test ({\tt test}) subsets, respectively.  We use the training subset to train our models, the development subset to select hyperparameters and the test subset to evaluate the final performance of the models.  Statistics for the {\it MIXED} and {\it COVID} datasets in terms of  number of tweets,  tokens, entities in the {\tt train}, {\tt test} and {\tt dev} subsets, respectively, are shown in Table \ref{train-test-dev}.
The benchmark datasets, together with the pre-processing script, will be made publicly available upon publication of this work. More specifically, to comply with Twitter's Developer Agreement and Policy\footnote{https://developer.twitter.com/en/developer-terms/agreement-and-policy}, the datasets will be made available as pairs of tweet ID and corresponding locations. The locations will be specified as a list of location-type tags corresponding to the tokens in the tweet as shown in Table \ref{examples} (i.e., a list of tags such as B-ctc, I-ctc, B-sta, O, etc. - one tag for each tweet token). Given that the pre-processing script will also be  made available, the index of the location tags should precisely match the index of the tweet tokens. In addition to the location annotated datasets of tweets, the IDs of the unlabeld tweets that are used to pre-trained BERT will also be made available for both mixes crises and COVID-19 health crisis.

\section{Background and Approaches}
\label{approaches}

The task of identifying fine-grained locations in tweet text can be formulated as follows: Given a set of $(X,Y)$ pairs, where $X=\{x_1,\cdots,x_n\}$ is a text sequence/tweet  with $n$ tokens, and $Y=\{y_1,\cdots,y_n\}$ is a tag sequence  with $n$ location tags/types (in BIO format) corresponding to the tokens in the sequence $X$; our sequence tagging task is to find a mapping $f_{\theta}:{\bf X}\rightarrow {\bf Y}$ (with parameters $\theta$) from input sequences to output sequences of fine-grained location types.

\subsection{Baseline Models}

\subsubsection{Feature-Engineered Baseline.}

\textbf{Stanford NER} \citep{Finkel:2005:INI:1219840.1219885} 
uses an arbitrary order linear chain CRF model over a set of predefined word and character level features extracted from the input. The model has been used as a strong baseline for many NER models. We retrain the model with both \textit{MIXED} and \textit{COVID} datasets, respectively, to learn  fine-grained location types. 

\subsubsection{Character and Word Embedding Baselines.}
One model architecture in this category consists of a distributed representation layer learning the embeddings at character and word level followed by an  LSTM-based context-encoder layer and a CRF tag-decoder. The model is referred as \textbf{CNN-GloVe-BiLSTM-CRF} in what follows.
Considering the recent success of transformer-based models, we also experiment with a similar model where BERT is used as the embedding layer instead of CNN+GloVe. We call this model \textbf{BERT-BiLSTM-CRF}.
For both CNN-GloVe-BiLSTM-CRF and BERT-BiLSTM-CRF models, we employ a multitask learning approach \cite{caruana1997multitask}, in which the main task of fine-grained location tagging is learned simultaneously with the auxiliary task of a  generic yes/no location tagging (see Appendix for more details).  We refer this model using the \textit{-MTL} suffix in what follows.

\subsubsection{Word and Entity Embedding Baseline.}
{In addition to using contextualized word embeddings learned from a transformer-based language model, \textbf{LUKE} \citep{yamada-etal-2020-luke} also learns contextualized entity embeddings and subsequently uses  an entity-aware self-attention mechanism to perform tasks such as entity typing, relation classification, NER, etc.  The LUKE approach has achieved state-of-the-art results on standard NER datasets (among others). We fine-tune the pre-trained \textit{LUKE-base} model with the \textit{COVID} and \textit{MIXED} datasets, respectively. The LUKE model selects candidate entity spans before making the entity type category predictions, a task that is comparable to the auxiliary task in the MTL models discussed earlier. Hence, we do not use the multitask learning setting for LUKE.

\section{Experimental Setup}
\label{setup}

In this section, we discuss the metrics used in the evaluation, implementation details and  experiments performed.

\subsection{Metrics} We use standard metrics, including  precision (Pr), recall (Re) and F1-measure (F1), to evaluate the performance of the models trained.

\subsection{Implementation details} 
We performed a grid-search with $5$ trials and used the development subsets to identify best-overall hyperparameter values (see the Appendix for details on the values included in the grid and best-overall values).
We used the best-overall values in the experiments.   We used the Glorot uniform initializer \citep{pmlr-v9-glorot10a} to initialize the model weights. The optimization was performed using the AdamW optimizer \citep{loshchilov2017decoupled}, with a learning rate of $1e-3$, weight decay of $1e-2$, and gradient clipping with max norm of $5$. We used a dropout of $0.5$ and mini-batch size of $32$ in all the experiments. 
We set a patience of $5$ epochs on the development F1-measure, as early stopping of training. All experiments are run on NVIDIA Tesla V100 GPU.\color{black}

\subsection{Experiments} We conducted experiments in two settings, {\it in-domain} and {\it cross-domain}.  In the {\bf in-domain setting}, models were trained and tested on the same dataset (e.g., models were trained on MIXED-train, tuned on MIXED-dev, and tested on MIXED-test).  The goal was to study: 1) the performance of the deep learning models by comparison  with the traditional Stanford NER model; 
2) the effect of the auxiliary task in the MTL framework; 3) the effect of different types of embeddings.  
In the {\bf cross-domain setting}, we used the best in-domain model to investigate several ways to perform transfer of information between domains: 1) a {\it zero-shot} transfer setting, where models trained on one dataset were tested on the other dataset (e.g., models trained on MIXED-train, tuned on MIXED-dev and tested on COVID-test); 2) an {\it embedding-level} transfer, where the transformer block fine-tuned on one dataset (e.g., MIXED) was used as a starting point for the transformer block of the model trained/tuned/tested on the other dataset (e.g., COVID); 3) a {\it model-level} transfer, where the  model trained/tuned on a dataset (e.g., MIXED-train, MIXED-dev) is used as the starting point of the model for the other dataset (e.g., COVID-train, COVID-dev, COVID-test, respectively). 

\section{Results and Discussion}
\label{results}

We first present and discuss the in-domain results, followed by the cross-domain results. In addition, we also perform error analysis and discuss the robustness of the models. 
\begin{table}[t]
\scriptsize
\centering
\begin{tabular}{|l|lll|}
\hline
\rowcolor{_blue!25!white} 
\textbf{Dataset}         & \multicolumn{3}{c|}{\textbf{MIXED}}                                                     \\ \hline
\textbf{Model}           & \multicolumn{1}{l}{\textbf{Pr}}    & \multicolumn{1}{l}{\textbf{Re}}    & \multicolumn{1}{l|}{\textbf{F1}}    \\ \hline
Stanford NLP (retrained) & \multicolumn{1}{l}{\textbf{82.52}} & \multicolumn{1}{l}{65.64}          & \multicolumn{1}{l|}{73.12}          \\
CNN-GloVe-BiLSTM-CRF     & \multicolumn{1}{l}{80.26} & \multicolumn{1}{l}{65.20}          & \multicolumn{1}{l|}{71.95}          \\
CNN-GloVe-BiLSTM-CRF-MTL & \multicolumn{1}{l}{76.92}            & \multicolumn{1}{l}{59.30}            & \multicolumn{1}{l|}{66.97}            \\
BERT-BiLSTM-CRF          & \multicolumn{1}{l}{74.98}          & \multicolumn{1}{l}{70.07}          & \multicolumn{1}{l|}{74.52}          \\
BERT-BiLSTM-CRF-MTL      & \multicolumn{1}{l}{74.58}          & \multicolumn{1}{l}{\textbf{74.75}} & \multicolumn{1}{l|}{74.67} \\
LUKE & 80.71 & 73.08 & \textbf{76.71} \\\hline
\rowcolor{_blue!25!white}
\textbf{Dataset}         & \multicolumn{3}{c|}{\textbf{COVID}}                                                     \\ \hline
{\bf Model}                   & \multicolumn{1}{l}{\textbf{Pr}}    & \multicolumn{1}{l}{\textbf{Re}}    & \multicolumn{1}{l|}{\textbf{F1}}    \\ \hline
Stanford NLP (retrained) & \textbf{85.71}                      & 57.62                               & 68.92                               \\
CNN-GloVe-BiLSTM-CRF     & 77.27                      & 68.81                               & 71.66                               \\
CNN-GloVe-BiLSTM-CRF-MTL & 78.64                                 & 52.74                                 & 63.14                                 \\
BERT-BiLSTM-CRF          & 73.41                               & 0.74                      & 72.05                               \\ 
BERT-BiLSTM-CRF-MTL      & \multicolumn{1}{l}{77.06}          & \multicolumn{1}{l}{69.43}          & \multicolumn{1}{l|}{73.04} \\
LUKE & 78.49 & \textbf{71.12 }& \textbf{74.66} \\\hline
\end{tabular}
\caption{In-domain results. Comparison of the following models: Stanford NER, CNN-GloVe-BiLSTM-CRF/BERT-BiLSTM-CRF and their MTL variants, and  LUKE.}
\label{in-domain}
\end{table}

\subsection{In-domain Setting}  Table \ref{in-domain} shows the in-domain results of the models.  As can be seen in Table \ref{in-domain},  the entity-embedding based LUKE model is the best overall in terms of F1-measure for both {\it MIXED} and {\it COVID} datasets, with a relatively high recall compared to most of the other models. Specifically, the F1-measure is 76.71\% for the {\it MIXED} dataset and 74.66\% for the {\it COVID} dataset. While the Stanford NLP has the highest precision overall, we argue that in the context of disaster monitoring and response, recall is more important than precision, as the final results will be reviewed by humans before any action is taken. Comparing the results for the {\it MIXED} and {\it COVID} datasets, we can see that the models  have slightly better performance on the {\it MIXED} dataset. While this dataset contains a variety of crisis events, the events are relatively localized to specific geographical regions, which may make it easier for the models to identify the locations. As opposed to that, the {\it COVID} dataset has a big variety of locations as it covers a global pandemic. Nevertheless, the F1 score of the LUKE model on {\it COVID}  is 8.3\% higher than the score of the Stanford NLP model, which uses manually designed features for training. We can also observe that the contextualized word and/or entity embeddings obtained from transformer architectures are better than both the engineered features in Stanford NLP and the character/word-embeddings in the CNN-GloVE-BiLSTM-CRF models. Finally, when comparing the BERT-BiLSTM-CRF-MTL model (with auxiliary task)  to its BERT-BiLSTM-CRF variant (without the auxiliary task), the results show that  the auxiliary task can help improve the F1-measure, especially in the case of {\it COVID}. However, for CNN-GloVe-BiLSTM-CRF, the addition of the auxiliary task decreases the F1-measure. This suggests that the transformer allows for a richer transfer of knowledge between similar tasks as compared to the CNN/GloVe architectures.
\begin{table}[t]
\scriptsize
\centering
\begin{tabular}{|l|l|lll|}
\hline
\rowcolor{_blue!25!white}
\multicolumn{2}{|c|}{\textbf{Datasets}} & \multicolumn{3}{l|}{\textbf{COVID$\rightarrow$MIXED}}        \\
\hline
\multicolumn{1}{|l|}{\textbf{Model}} &
\multicolumn{1}{l|}{\textbf{Transfer style}} & \textbf{Pr}    & \textbf{Re}    & \textbf{F1}    \\
\hline
\multirow{3}{*}{BERT-BiLSTM-CRF-MTL} &
\multicolumn{1}{l|}{zero-shot}         & 77.05          & 54.80          & 64.05          \\
&\multicolumn{1}{l|}{embedding-level}   & 76.40          & 71.20 & 73.71          \\
&\multicolumn{1}{l|}{model-level}       & 79.86 & 71.05          & 75.20 \\ \hline
\multirow{3}{*}{LUKE} &
zero-shot                               &     50.65      & 47.41          &     48.98      \\ 
&embedding-level                         & 78.13 & \textbf{74.21} & 76.12           \\ 
&\multicolumn{1}{l|}{model-level}       & \textbf{81.32} & 73.57 & \textbf{77.25} \\ \hline
\rowcolor{_blue!25!white}
\multicolumn{2}{|c|}{\textbf{Datasets}} & \multicolumn{3}{l|}{\textbf{MIXED$\rightarrow$COVID}}        \\ \hline
\multicolumn{1}{|l|}{\textbf{Model}} &
\multicolumn{1}{l|}{\textbf{Transfer style}} & \textbf{Pr}    & \textbf{Re}    & \textbf{F1}    \\ \hline
\multirow{3}{*}{BERT-BiLSTM-CRF-MTL} &
zero-shot                               & 36.23          & 45.85          & 40.41          \\ 
&embedding-level                         & 67.19 & 68.85          & 68.01          \\ 
&\multicolumn{1}{l|}{model-level}       & 66.44          & 71.47 & 68.86 \\ \hline
\multirow{3}{*}{LUKE} &
zero-shot                               &     \textbf{79.94}      & 43.17          &      56.06     \\ 
&embedding-level                         & 76.95 & \textbf{73.78} & 75.33           \\ 
&\multicolumn{1}{l|}{model-level}      & 78.08 & 73.32 & \textbf{75.63}  \\ \hline
\end{tabular}
\caption{Cross-domain results. Comparison between three transfer styles: zero-shot, embedding-level and model-level. }
\label{cross-domain}

\end{table}

\subsection{Cross-domain setting} Table \ref{cross-domain} shows the results of the  BERT-BiLSTM-CRF-MTL and LUKE models (which give the best overall results in the in-domain setting) in the cross-domain setting. Specifically, we compare three transfer styles, {\it zero shot}, {\it embedding-level}, and {\it model-level}, when {\it COVID} is used as source and {\it MIXED} as target, and the other way around. As expected, the  model-level transfer style gives the best results overall, while the  zero-shot style gives the worst results overall.
Notably, in the case of the {\it COVID} to {\it MIXED} transfer,  the model-level transfer improves the results of the in-domain LUKE model, from 76.71\% to 77.25\%. This is probably due to the diversity in the {\it COVID} dataset, which enables more accurate locations to be identified in the {\it MIXED} dataset. As opposed to that, the transfer from {\it MIXED} to {\it COVID} causes more specific locations to be identified, which improves the recall but  negatively affects the precision (and the overall F1-measure).

\subsection{Error analysis}
We performed error analysis of the model-level transfer from Table \ref{cross-domain} for both BERT-BiLSTM-CRF-MTL and LUKE (specifically, model-level transfer from {\it COVID} to {\it MIXED} and  from {\it MIXED} to {\it COVID}). The analysis is based on the framework proposed by \citet{ribeiro-etal-2020-beyond}, where a model is tested for a capability using three tests: minimum functionality test (MFT), invariance test (INV) and directional expectation test (DIR). We performed the tests on the model's capability to generalize the concept of a location entity. In our case, MFT is the model's performance on the original {\it MIXED} or {\it COVID} test set, respectively. For INV, the location entities in the original test set were replaced with other randomly selected location entities of the same type from the test set. Finally, for DIR, the original location entities were replaced with randomly selected location entities of different types from the test set. 
The results of the analysis are shown in Table \ref{error_analysis}. The MFT score serves as a baseline for the other two tests. As can be seen, in both cases, the performance degrades when the locations are mixed up - tests INV and DIR as compared with the test MFT - suggesting that the model captures correlations between locations and their context. 
However, the F1 score for INV is better than the F1 score for DIR, which shows that the model expects a particular type of location in a given context. 

Table \ref{error_analysis_examples} shows sample predictions for different tests (MFT, INV, DIR). In the first example,  for the MFT test, the model makes a correct prediction for a tweet where a location entity of type \textit{ctc} is followed by a location entity of type \textit{sta}, which is the general convention for specifying a {\it city, state} location. However, for the DIR test, when the entities are replaced with others in reverse order of the type  as compared to the original tweet (i.e., {\it sta, ctc} instead of {\it ctc, sta}), 
the model incorrectly, but not surprisingly, predicts \textit{sta} as \textit{ctc} and vice versa. In the second example, for the MFT test, the model correctly predicts Sri Lanka as a country (i.e., {\it con}). However, when \textit{Sri Lanka} is replaced with \textit{South Africa} in the case of the INV test, the model predicts it as \textit{reg}. This is probably because Africa as a continent is a location of type \textit{reg}, and also because cardinal directions are commonly associated with  \textit{reg} locations. Hence, without any external knowledge about \textit{South Africa} as a country, \textit{reg} is the next best prediction.
\begin{table}[t]
\centering
\scriptsize
\begin{tabular}{|c|l|l|l|l|}
\hline
\rowcolor{_blue!25!white}
 \multicolumn{5}{|c|}{\textbf{Model level COVID$\rightarrow$MIXED transfer}}        \\
\hline
\textbf{Model}&\textbf{Test} & \textbf{Pr} & \textbf{Re} & \textbf{F1} \\ \hline
\multirow{3}{*}{BERT-BiLSTM-CRF-MTL} & MFT & \textbf{79.86} &  \textbf{71.05} & \textbf{75.20}\\\cline{2-5}
& INV & 67.78 & 52.29 & 59.03 \\ \cline{2-5}
& DIR & 47.48 & 31.34 & 37.76 \\ \hline
\multirow{3}{*}{LUKE} & MFT & \textbf{81.32} &  \textbf{73.57} & \textbf{77.25}\\\cline{2-5}
& INV & 70.52 & 50.71 & 58.99 \\ \cline{2-5}
& DIR & 56.87 & 34.41 & 42.88 \\ \hline
\rowcolor{_blue!25!white}
 \multicolumn{5}{|c|}{\textbf{Model level MIXED$\rightarrow$COVID transfer}}         \\
\hline
\multirow{3}{*}{BERT-BiLSTM-CRF-MTL} & MFT & \textbf{66.44} &  \textbf{71.47} & \textbf{68.86}\\\cline{2-5}
& INV & 57.64 & 59.35 & 58.48 \\ \cline{2-5}
& DIR & 40.26 & 33.80 & 36.75 \\ \hline
\multirow{3}{*}{LUKE} & MFT & \textbf{78.08} &  \textbf{73.32} & \textbf{75.63}\\\cline{2-5}
& INV & 69.86 & 53.43 & 60.55 \\ \cline{2-5}
& DIR & 49.47 & 29.28 & 36.79 \\ \hline
 
\end{tabular}
\caption{Error analysis tests (MFT, INV and DIR) for the capability of the model-level transfer approach to generalize the concept of a  location entity.}
\label{error_analysis}
\end{table}

\section{Conclusions and Future Work}
\label{conclusions}

In this paper, we introduced two new crisis tweet datasets manually tagged with specific fine-grained location types. These are the first manually annotated datasets for fine-grained location identification in crisis tweet texts, and can foster research in this area of great importance for crisis monitoring and response. The two datasets are different in nature, with one of them focused on mixed natural and man-made crisis events, which are generally localized to specific regions, and the second one focused on the worldwide COVID-19 pandemic. The different nature of the two datasets  enables studies on location identification for localized and global events, as well as studies on the transferability of information between localized  and global events.

In addition to introducing these datasets, we reported baseline results for the fine-grain location identification task using state-of-the-art models based on different embedding styles. Our results suggest that the entity-embedding style of the LUKE model gives the best results. We also used MTL to incorporate an auxiliary task in one of the models and showed its effectiveness in transferring information between datasets. As part of future work, we plan to improve the results of the models by including other crisis-related tagging and classification tasks in the LUKE/MTL settings.
\begin{table}[t]
\scriptsize
\centering
\begin{tabular}{|c|l|l|}
\hline
\multicolumn{1}{|l|}{{\bf No.}} &{\bf Test} &{\bf Tweet text}                                                                                                      \\ \hline

 \multirow{5}{*}{1} & \multirow{3}{*}{MFT}  & NEW Hurricane Florence has made landfall near  
\\
                      & &  \locationtag[ctc$\rightarrow$ctc]{Wrightsville Beach,} \locationtag[sta$\rightarrow$sta]{North Carolina}\\
                    & & at 7 15 AM EDT\\
                    \cline{2-3}
                &\multirow{2}{*}{DIR}  & NEW Hurricane Florence has made landfall near \\ 
                & & \locationtag[sta$\rightarrow$ctc]{WI,} \locationtag[ctc$\rightarrow$sta]{Panama} at 7 15 AM EDT\\%
                    
                          \hline 
\multirow{6}{*}{2} & \multirow{3}{*}{MFT}  & On this Easter Sunday my thoughts are with \\
& & \locationtag[con$\rightarrow$con]{Sri Lanka} following the horrific attacks
\\
                    & &  on worshippers there.\\
                    \cline{2-3}
                & \multirow{3}{*}{INV}  & On this Easter Sunday my thoughts are with \\
                & & \locationtag[con$\rightarrow$reg]{South Africa} following the horrific attacks on
\\
                    & &  worshippers there.\\
                    
                          \hline

\end{tabular}
\caption{Examples of location predictions for different error analysis test settings (MFT, INV and DIR). The examples are from the {\it MIXED} datasets and the predictions are made using  model-level transfer from {\it COVID} to {\it MIXED}. The locations are highlighted with pink. The labels and predictions for   entities are shown as a subscript to the corresponding locations using the convention \textit{gold label$\rightarrow$model prediction}.}
\label{error_analysis_examples}
\end{table}

\section{Ethics and Impact Statement}
\label{ethics}
The  dataset that we plan to share will not provide any personally identifiable information, as only the tweet IDs and human annotated location tags (i.e., tags such as B-ctc, I-sta, O, etc., but  not specific locations) will be shared. Thus, our dataset complies with
Twitter's Developer Agreement and Policy\footnote{https://developer.twitter.com/en/developer-terms/agreement-and-policy} in terms of privacy. 
Furthermore, in compliance with the Twitter's Developer Agreement and Policy, Section III.E, the location information is used only in conjunction with the tweet content, and, as allowed by Twitter, we {\tt "only use such location data and geographic information to identify the location tagged by the Twitter Content."} In terms of impact, the research enabled by this dataset has the potential to help officials and health organizations identify actionable information useful for fast response during a crisis situation, or  facilitate the health organizations to aggregate information relevant to COVID-19 by locations (which in turn can be useful in  preventing a serious resurgence of the novel coronavirus in a particular region). However, we want to emphasize that we do not use any of the information in Twitter content, in particular the location information, to infer  any sensitive information about the user, and most importantly our models  do not infer any information about users' health\footnote{https://developer.twitter.com/en/developer-terms/more-on-restricted-use-cases}. The models are simply trained to identify location tags in tweets (as explicitly allowed by Twitter) and nothing more. Also important, our pre-processing script removes any user mentions from the tweet content before feeding the tweets to the models for training.

\bibliography{aaai22.bib}

\end{document}


\appendix
\begin{table*}[h]
\small
\centering
\begin{tabular}{|l|l|l|l|}
\hline
\textbf{Dataset }               & \textbf{Event}                                                                                                        & \textbf{Keywords} & \textbf{Size} \\
 \hline  
\multirow{5}{*}{MIXED} & \begin{tabular}[c]{@{}l@{}}Nepal Earthquake \\ and Queensland Floods\\
(Alam, Joty, and Imran 2018) 
\end{tabular} &    N/A       &    167  \\ \cline{2-4}
                       &                                              
                       \begin{tabular}[c]{@{}l@{}}Srilanka Bombing\\ (ours)      
                       \end{tabular}
                       & \begin{tabular}[c]{@{}l@{}}Sri lanka attack, Sri lanka terror, Sri Lanka horror, Sri Lanka easter
                       \end{tabular}
                       &   1171   \\ \cline{2-4}
                       & 
                 \begin{tabular}[c]{@{}l@{}}Hurricane Michael\\ and Hurricane Florence (ours)      
                       \end{tabular}
                       & \begin{tabular}[c]{@{}l@{}}
                            hurricane michael, hurricanemichael, hurricane florence, hurricaneflorence
                       \end{tabular}
                       &    2758  \\     \hline
COVID                  & 

                        \begin{tabular}[c]{@{}l@{}}COVID-19 (ours)      
                       \end{tabular}  
& \begin{tabular}[c]{@{}l@{}}
\#coronavirus, corona virus, \#Coronavid19, \#coronavirususa, \#covid19,\\\#covid-19, \#coronapocalypse, \#quarantinelife, \#socialdistancing
\end{tabular}&   5243  \\ \hline
\end{tabular}
\caption{Keywords used to collect tweets and the number of tweets from each event in the {\it MIXED} and {\it COVID} datasets.}
\label{events}
\end{table*}

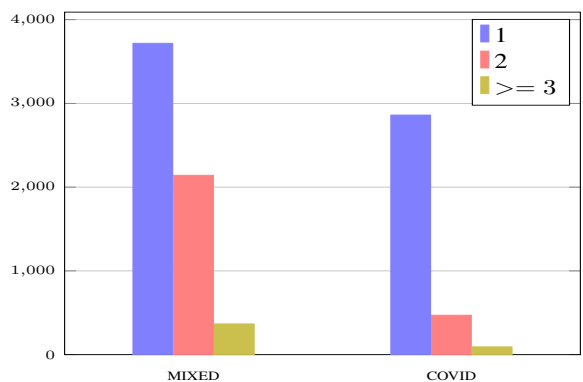
\begin{figure*}[h]
\centering
\begin{tikzpicture}[trim axis right,yscale=0.8]
    \begin{axis}[
        major x tick style = transparent,
        ybar=1*\pgflinewidth,
        bar width=15pt,
        ymajorgrids = true,
        symbolic x coords={MIXED,COVID},
        xtick = data,
        scaled y ticks = false,
        enlarge x limits=0.5,
        ymin=0,
        legend cell align=left,
    ]
\addplot[style={blue!50!white,fill=blue!50!white,mark=none}] coordinates
{(COVID,2861) (MIXED,3718)};
\addplot[style={red!50!white,fill=red!50!white,mark=none}] coordinates
{(COVID,471) (MIXED,2142)};
\addplot[style={yellow!75!black,fill=yellow!75!black,mark=none}] coordinates
{(COVID,93) (MIXED,367)};
        \legend{1,2,$>=3$}
    \end{axis}
\end{tikzpicture}
\caption{Entity distribution by number of tokens in a location in the {\it MIXED} and {\it COVID} datasets, respectively.}
\label{fig:tokens_per_entity}
\end{figure*}

\begin{figure*}[h]
  \centering
  \includegraphics[width=25em]{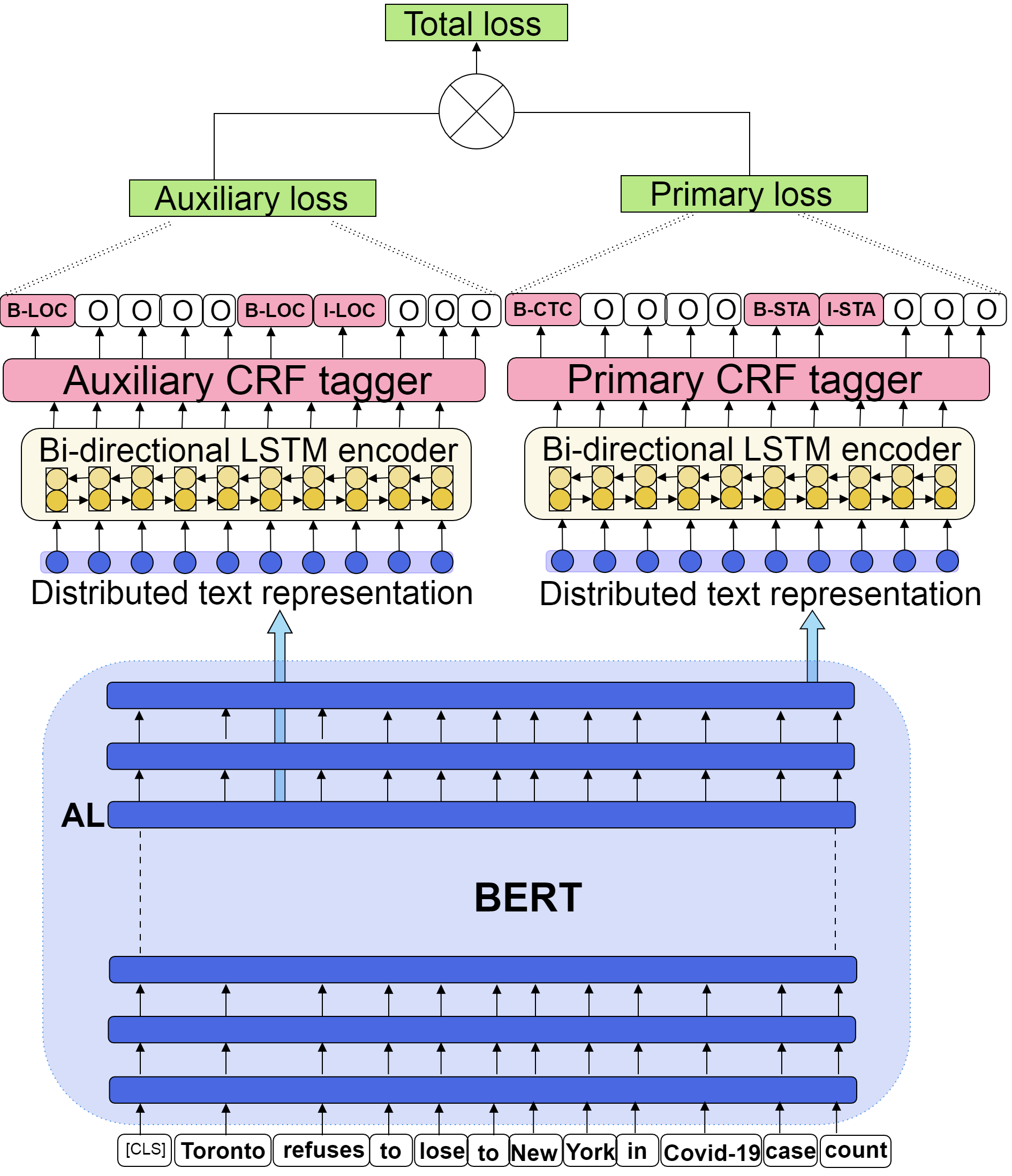}
    \caption{BERT-BiLSTM-CRF based MTL model. The model can be seen as an MTL model, with two objective corresponding to two tasks. The primary task (right) is to predict fine-grained location tags, while the auxiliary task (left) is to predict generic location tags. BERT is used to get a distributed representation of the input for both models. The primary task is linked to the last BERT layer, while the auxiliary task is linked to a lower layer (AL). BiLSTM and CRF models are used as context encoders and tag decoders for both tasks.} 
  \label{model_architecture}
\end{figure*}

\begin{table*}[h]
\centering
\begin{tabular}{|l|l|}
\hline
\textbf{Hyperparameters} & \textbf{Search space} \\
\hline
Fine-tuned BERT layers & 9, 10, \textbf{11}\\ 
Auxiliary task  layer (AL)  & None, 6, 7, \textbf{8}, 9, 10\\  
Auxiliary loss factor & 0.2 \\
Hidden size of BiLSTM & 64, 128, 256, {\bf 512}\\
\hline
\end{tabular}
\caption{Hyperparameter search space for the BERT-BiLSTM-CRF model. The best overall values based on the development subset are highlighted with bold font. 
}
\label{hyperparameters}
\end{table*}